%% file: iclr2023_conference.tex
\documentclass{article} 
\usepackage{iclr2023_conference,times}

\input{math_commands.tex}

\usepackage{hyperref}
\usepackage{url}
\usepackage{multirow}
\usepackage{arydshln}
\usepackage{graphicx}
\usepackage{threeparttable}

\usepackage{kotex}

\title{D-TensoRF: Tensorial Radiance Fields for Dynamic Scenes}

\author{Hankyu Jang \& Daeyoung Kim 
\\
School of Computing\\
KAIST\\
Daejeon, South Korea \\
\texttt{\{gksrb97, kimd\}@kaist.ac.kr} \\
}

\iclrfinalcopy 
\begin{document}

\maketitle

\begin{abstract}
 Neural radiance field (NeRF) attracts attention as a promising approach to reconstructing the 3D scene. As NeRF emerges, subsequent studies have been conducted to model dynamic scenes, which include motions or topological changes. However, most of them use an additional deformation network, slowing down the training and rendering speed. Tensorial radiance field (TensoRF) recently shows its potential for fast, high-quality reconstruction with compact model size by using multiple factorized components of an explicit data structure (4D tensor) and a small-sized neural network. Nonetheless, TensoRF is still only applicable to static scenes. In this paper, we present D-TensoRF, a tensorial radiance field for dynamic scenes, enabling novel view synthesis at a specific time. We consider the radiance field of a dynamic scene as a 5D tensor. The 5D tensor represents a 4D grid in which each axis corresponds to X, Y, Z, and time and has multi-channel features per element. Similar to TensoRF, we decompose the grid either into rank-one vector components (CP decomposition) or low-rank matrix components (MM decomposition). We newly propose matrix-matrix (MM) decomposition to reduce the amount of computation compared to CP decomposition and shorten the training time. We also use smoothing regularization to reflect the relationship between features at different times (temporal dependency). Through ablation study, we reveal that smoothing regularization is essential when modeling dynamic scenes. Above all, D-TensoRF with both CP and MM decomposition have significantly more compact model sizes than existing works. We show that D-TensoRF with CP decomposition and MM decomposition both have short training times and low memory footprints with quantitatively and qualitatively competitive rendering results in comparison to the state-of-the-art methods in 3D dynamic scene modeling.

\end{abstract}

\section{Introduction}
The reconstruction and modeling of 3D scenes play an essential role in various applications such as virtual reality, 3D visual content creation, and the film and game industry. Neural radiance fields (NeRF)(\cite{mildenhall2021nerf}) and recent subsequent research (\cite{TensoRF}, \cite{mueller2022instant}) based on it achieve photo-realistic rendering.

Recently, methods using explicit data structures(\cite{Plenoctree}, \cite{baking_nerf}, \cite{Directvoxgo}, \cite{mueller2022instant}, \cite{TensoRF}) show faster training and rendering speeds. Tensorial radiance fields (TensoRF)(\cite{TensoRF}) accomplishes a fast training time by using a 3D feature grid and small-sized of a neural network simultaneously. TensoRF even decomposes the feature grid into low-rank tensor components to achieve a lower memory footprint than other methods using the explicit data structure.

Most NeRF methods focus on modeling static scenes rather than dynamic scenes with movement. However, the 3d reconstruction from the images of a moving object is much more applicable to various real-life applications. Especially, modeling a dynamic scene is also indispensable when considering the data collection process in real life. In the data collection process, an object is photographed simultaneously with a large number of cameras or photographed multiple times from several directions with a small number of cameras. The latter option is commonly used due to affordable expenses. In this case, smoothly changing images are obtained according to the shooting time because many real-life cases include object motions or topological changes. 

Including D-NeRF(\cite{pumarola2020d}), many studies(\cite{hypernerf}, \cite{DBLP:journals/corr/abs-2012-12247}, \cite{li2020neural}) have been conducted recently to model a dynamic scene. Most studies use an additional deformation network that maps the coordinates of the sampled points to the canonical space.  Since the deformation network is the same size as the radiance network that predicts color and density, it involves a longer training time and slower rendering speed.

Inspired by the idea of TensoRF, we represent the dynamic scene as a 5D tensor which represents a 4D grid in which each axis corresponds to X, Y, Z, and time. Moreover, each element of the 4D grid has a single-channel or multi-channel feature. Then, we applied tensor decomposition to the 4D grid.

We attempt the classic CANDECOMP/PARAFAC (CP) decomposition(\cite{Carroll1970}) to factorize the 5D tensor first. Even with CP decomposition, we produce competitive rendering results compared to the state-of-the-art methods. Even though CP decomposition makes the model substantially compact, more components are required to model complex scenes, which increases computational cost and training time. To prevent this limitation, we propose a matrix-matrix (MM) decomposition that factorizes the full tensor of a radiance field into multiple matrix factors. Although the model size is slightly increased in contrast to when using the CP decomposition, the computational overhead is reduced significantly.

However, naively extending the time dimension of the TensoRF framework does not produce the desired result. This is because it does not consider the temporal dependency, which is the relation between the scenes at different times.  According to our observation that the scenes at the adjacent times are closely related, we apply a smoothing regularization along the time axis to reflect the temporal dependency. 

We conduct the experiments with our models (both with CP decomposition and MM decomposition) in various settings (different grid resolutions and the number of components). We show that our model's rendering results are competitive compared to existing methods and demonstrate efficiency in terms of memory footprint and training speed. Our model with MM decomposition achieves satisfactory performance with only 8 minutes of training, and the model with CP decomposition requires only 1.8 MB. In addition, we carry out an ablation study on smoothing regularization to prove its importance.

To sum up, our main contributions are listed as follows:

\begin{itemize}
    \item We express a dynamic scene with object motion or topological change from a tensorial perspective. We suggest a solution to model the dynamic scene by optimizing low-rank tensor components. We factorize the low-rank tensor components with the CP/MM decomposition and each component has factors related to the X, Y, Z, and time axes.
    \item We apply the smoothing regularization to the time-related factor of each low-rank tensor component to reflect the temporal dependency.
    \item Both our models with CP decomposition and MM decomposition achieve good rendering quality in both qualitative and quantitative terms. More importantly, our models are significantly compact and fast in training.
\end{itemize}

\section{Related Work}

\subsection{Neural Radiance Fields}

Different kinds of 3D scene representations, such as meshes, point clouds, volumes, and implicit representations(\cite{Learning_implicit_fields}, \cite{Lombardi:2019}, \cite{mildenhall2019llff}, \cite{extracting_triangular}, \cite{Deepsdf}, \cite{sitzmann2019srns}, \cite{mesh_reconsturction}), have been studied in order to improve rendering quality. To represent a 3D scene, NeRF(\cite{mildenhall2021nerf}) proposes to use a neural network as a function that receives position and viewing direction and outputs corresponding color and density. This representation achieves photo-realistic results and shows its potential. It has been studied extensively and applied in various graphic or vision applications such as surface reconstruction(\cite{UNISURF}), fast rendering(\cite{fastnerf}, \cite{kilonerf}), texture mapping(\cite{baatz2021nerftex}), and dynamic scenes(\cite{pumarola2020d}, \cite{hypernerf}, \cite{park2021nerfies}, \cite{tineuvox}).

Recent approaches to this representation can be categorized into implicit, explicit, and hybrid representations. The implicit representation, such as NeRF(\cite{mildenhall2021nerf}) is purely MLP-based. Even though the representation is straightforward, the high computation cost of MLP impedes training and rendering speed. The explicit representation is the method that tries to model a 3D scene with only explicit data structures storing features, not using a single neural network. Plenoxels(\cite{plenoxel}) only uses a sparse 3D voxel grid with spherical harmonics and achieves desirable rendering quality. Although it is possible to reduce the expensive computation of MLP, the memory footprint is increased compared to the implicit representation. Hybrid representations use both an explicit data structure (sparse voxel grids(\cite{liu2020neural}), octrees(\cite{Plenoctree}), multi-resolution hashmap(\cite{mueller2022instant}), low-rank tensor components(\cite{TensoRF})) and a small neural network. This lessens the size of the model and training time concurrently.  

Several types of research to model the radiance field for dynamic scenes have also been conducted. One is to extend the time dimension in the radiance field for a static scene(\cite{du2021nerflow}, \cite{DBLP:journals/corr/abs-2105-06468}, \cite{pumarola2020d}). Most approaches take advantage of the additional deformation field to predict motions(\cite{pumarola2020d}, \cite{hypernerf}, \cite{tineuvox}). The deformation field maps the point coordinates into the canonical space and passes them to the radiance network as inputs. To improve rendering quality, HyperNeRF(\cite{hypernerf}) uses hyperspace to reflect the discontinuity of the deformation field, and 
NR-NeRF\cite{DBLP:journals/corr/abs-2012-12247} segments a nonrigid foreground and a rigid background. However, the previous methods are based on implicit representations, which are very slow in training. Recently, TiNeuVox(\cite{tineuvox}) accelerates the training speed by using explicit voxel features with neural networks, including the deformation and radiance network. 

Our model is a kind of hybrid representation. We utilize both an explicit 4D grid and a radiance network. Unlike the conventional methods, we do not use any other deformation network.

\subsection{Tensor Decomposition}
Tensor decomposition is one of the effective methods for analyzing tensors, which are multi-dimensional arrays. Tensor decomposition of high-order tensors(\cite{Tensor_decomposition}) can be thought of as an extended, generalized version of matrix singular value decomposition.
Various fields mainly use Tucker decomposition(\cite{Tucker1966}) and CP decomposition(\cite{Carroll1970}). Tucker decomposition factorizes a tensor into a core tensor multiplied by a matrix along each mode. CP decomposition is a particular case of Tucker decomposition in which a core tensor is restricted to being diagonal.

Tensor decomposition has been used in many fields, for example, anomaly detection(\cite{4781131}), recommender systems(\cite{SymeonidisZioupos16}), or multiple vision tasks(\cite{Block_term_NN}, \cite{compact_rnn}). TATD (Time Aware Tensor Decomposition)(\cite{tatd}) is suggested to decompose temporal tensors for missing entry prediction by considering the characteristics of temporal tensors: temporal dependency and time-dependent sparsity. 

In particular, TensoRF(\cite{TensoRF}) decomposes the neural radiance field using tensor decomposition(CP decomposition and vector-matrix decomposition (VM) for the first time. TensoRF, thereby, successfully reduces the training time and memory costs for storing features. 

Moreover, CCNeRF(\cite{tang2022compressible}) only uses pure tensor rank decomposition without using MLP. This method enables compression of the neural radiance field by maintaining low-rank approximation properties. 

We consider the dynamic radiance field as a temporal tensor, including the time axis, and factorize it through tensor decomposition. Also, we apply the smoothing regularization for temporal dependency.

\section{Preliminary}\label{Preliminary}

\subsection{Neural Radiance Fields}

NeRF (\cite{mildenhall2021nerf}) is a representation of a 3D volumetric scene. This representation approximates the scene with a function $F_{\Theta }:(\textbf{x}, \textbf{d})\to (\textbf{c}, \sigma)$. The function is simply implemented with MLP. $\textbf{x}$ denotes 3d coordinates of points $(x,y,z)$ and $\textbf{d}$ is 2d viewing direction $(\theta, \phi)$. Each $\textbf{c}, \sigma$ corresponds to emitted color and volume density. 
Given a camera ray $\textbf{r}$ whose origin is $\textbf{o}$ and direction is $\textbf{d}$  can be expressed as  $\textbf{r} = \textbf{o} + t\textbf{d} $ where $t$ is distance from predefined range $[t_n, t_f]$. Points sampled along ray $\mathbf{x_i}= \mathbf{o} + t_i\mathbf{ d}$ are taken as inputs of $F_{\Theta}$. The function $F_{\Theta}$ outputs colors $c_i$ and densities $\sigma_i$. The expected color of the pixel corresponding to the ray can be estimated by the quadrature rule:
\begin{align}\label{eq:1}
    \hat{\textbf{C}}(r) = \sum_{i=1}^{N}T_i(1-exp(-\sigma_i\delta _i))\textbf{c}_i, T_i =exp(-\sum_{j=1}^{i-1}\sigma_j\delta_j)
\end{align}

where $\delta_i = t_{i+1}- t_i$ is a distance between sampled points. Finally, NeRF is optimized with 2D images from different viewing directions by minimizing the following loss:

\begin{align}\label{eq:2}
    \mathit{L}=\left\|\hat{\textbf{C}}(\textbf{r})-\textbf{C}(\textbf{r}) \right\|^2
\end{align}

where $\textbf{C}(\textbf{r})$ is ground truth color of the pixel.

\subsection{Tensorial Radiance Fields} 

The original NeRF (\cite{mildenhall2021nerf}) is computationally expensive because it uses 8 fully-connected layers. Therefore, many works tried to reduce the computational cost by adopting an explicit data structure. However, explicit data structure requires much memory to store features. Tensorial Radiance Fields (TensoRF)(\cite{TensoRF}) is proposed while pointing out these two limitations of existing methods. 

TensoRF uses two 3D voxel grids to retain color-related features(multi-channel) or density-related features(single channel). After obtaining features according to the 3d location $\textbf{x}$, the density-related feature values are used directly as volume density $\sigma$. The color-related feature vectors are converted into view-dependent color $\textbf{c}$  under the condition of the viewing direction $\textbf{d}$ through the pre-defined function $S$. This grid based-radiance field can be expressed as follows:

\begin{align}\label{eq:3}
    \sigma, c = \mathbf{G}_\sigma(\textbf{x}), S(\mathbf{G}_c(\textbf{x}),\textbf{d})
\end{align}

Geometry grid(density-related) is a 3D tensor $\mathbf{G}_{\sigma}\in \mathbb{R}^{I \times J \times K}$ and appearance grid(color-related) is a 4D tensor $\mathbf{G}_{c}\in \mathbb{R}^{I \times J \times K \times P}$. $I$, $J$, and $K$ correspond to the resolution of the feature grid along the $X$, $Y$, and $Z$ axes, respectively, and $P$ is the number of feature channels. 

TensoRF factorizes the geometry grid $\mathbf{G}_{\sigma}$ and the appearance grid $\mathbf{G}_{c}$ in two ways to make the model size compact. 
First, it directly applies the classic CP decomposition, which factorizes the tensor as a summation of rank-one tensor components, where each component is the outer product of the vector factors. With CP decomposition, Geometry grid $\mathbf{G}_\sigma$ and appearance grid $\mathbf{G}_c$ are decomposed as:

\begin{align}\label{eq:4}
\mathbf{G}_\sigma = \sum_{r=1}^{R_\sigma}v_{\sigma,r}^{X}\circ v_{\sigma,r}^{Y} \circ v_{\sigma,r}^{Z}
\end{align}

\begin{align}\label{eq:5}
\mathbf{G}_c = \sum_{r=1}^{R_c}v_{c,r}^{X}\circ v_{c,r}^{Y} \circ v_{c,r}^{Z} \circ b_r
\end{align}

$v^X$, $v^Y$, and $v^Z$ denote the vector factors associated with $X$,$Y$ and $Z$ axis and $b_r$ is an additional vector for feature channel dimension.
Due to the high compactness of CP decomposition, it requires a large number of $R_\sigma, R_c$ to model complex 3D scenes. This leads to an increase in the computational cost

Second, TensoRF proposes using Vector-Matrix (VM) decomposition, which factorizes the tensor into multiple vectors and matrices as follows:
\begin{align}\label{eq:6}
    \mathbf{G}_\sigma = \sum_{r=1}^{R_\sigma}v_{\sigma,r}^{X}\circ M_{\sigma,r}^{Y,Z} + v_{\sigma,r}^{Y}\circ M_{\sigma,r}^{X,Z} + v_{\sigma,r}^{Z}\circ M_{\sigma,r}^{X,Y}
\end{align}

\begin{align}\label{eq:7}
    \mathbf{G}_c = \sum_{r=1}^{R_c}v_{c,r}^{X}\circ M_{c,r}^{Y,Z}\circ b_{3r-2} + v_{c,r}^{Y}\circ M_{c,r}^{X,Z}\circ b_{3r-1} + v_{c,r}^{Z}\circ M_{c,r}^{X,Y}\circ b_{3r}
\end{align}

Matrix factors include information corresponding to two of the $X$, $Y$, and $Z$ axes. 
Compared to using CP decomposition, TensoRF with VM decomposition requires more memory because matrices have more parameters than vectors. However, since matrices have much more information, TensoRF with VM decomposition achieves high rendering quality with only a small number of components, effectively reducing the computational cost. 
Finally, it renders images with a combination of Eqn.\ref{eq:1}, \ref{eq:3}

\section{Method}

In section \ref{DTensoRF}, \ref{Factorization}, we explain how the dynamic radiance field is represented and factorized from a tensorial perspective. Then we describe how the smoothing regularization is applied in section \ref{Smoothing regularization}. Finally, the implementation details are illustrated in section \ref{implementation_details}.

\subsection{Tensorial radiance field representation for dynamic scene} \label{DTensoRF}

TensoRF utilizes multiple photos of a non-moving object, but the data we can easily obtain in the real world often involves movement over time. Our model aims to optimize the dynamic radiance field, which is a function $F$ that maps 3D location $\textbf{x} = (x,y,z)$, viewing direction $\textbf{d}=(\theta, \phi)$ and time instant $t$ to emitted color $\textbf{c}$ and volume density $\sigma$. This can be written as follows:
\begin{align}\label{eq:8}
    F: (\textbf{x}, \textbf{d}, t) \to (\textbf{c}, \sigma)
\end{align}

Similar to TensoRF, we model the function $F$ with 4D grids, 1) a geometry grid $\mathbf{G}_\sigma$ and 2) a appearance grid $\mathbf{G}_c$. Each dimension of the 4D grid corresponds to the $X$, $Y$, $Z$, and time axis and stores features in each element of the grids. Our model outputs emitted color $\textbf{c}$ from the appearance feature and the viewing direction $\textbf{d}$ with an extra converting function $S$. We use the value of the geometry feature as volume density $\sigma$. This dynamic tensorial radiance field(D-TensoRF) can be rewritten as:
\begin{align}\label{eq:9}
    \sigma, c = \mathbf{G}_\sigma(\textbf{x}, t), S(\mathbf{G}_c(\textbf{x},t),\textbf{d})
\end{align}

$\mathbf{G}_\sigma(\textbf{x}, t), \mathbf{G}_c(\textbf{x},t)$ are quadrilinear interpolated features from the geometry grid and the appearance grid respectively at location $\textbf{x}$ and at time $t$. We can use both a small-sized MLP(2 fully-connected layers) and spherical harmonics as $S$, but MLP consistently shows better results.  

\subsection{Factorizing Dynamic Radiance Field} \label{Factorization}

\begin{figure}[h] 
\begin{center}

\centerline{\includegraphics[width=15cm]{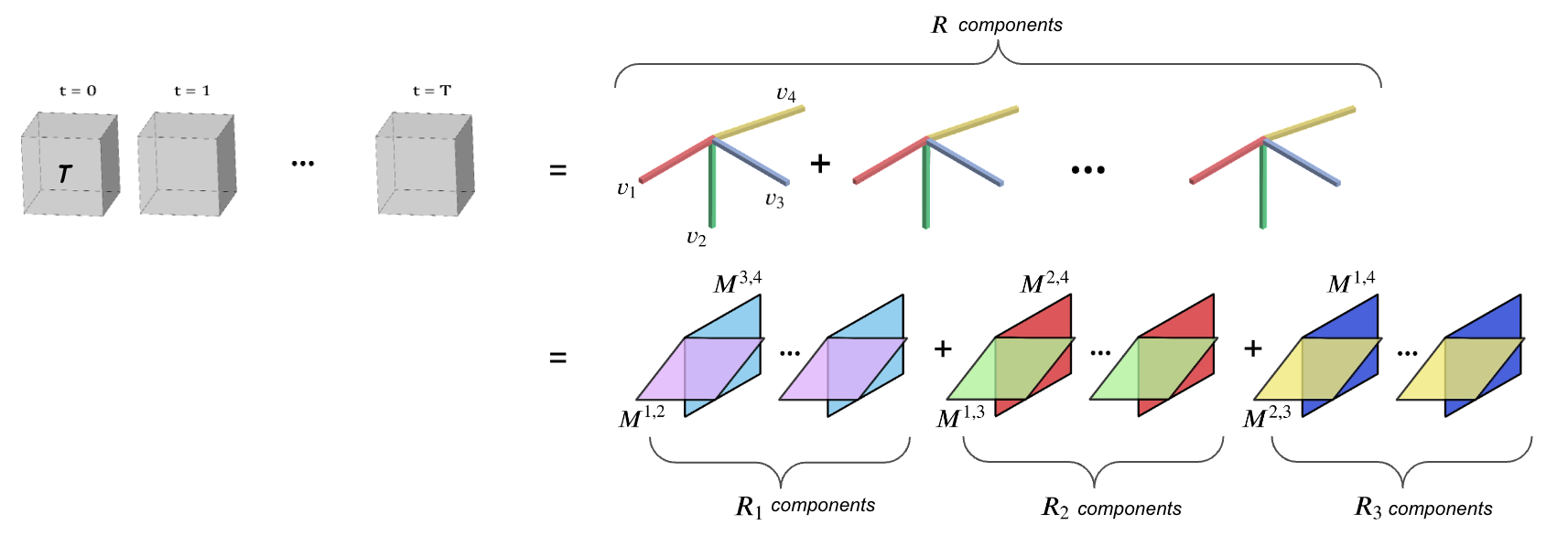} }

\end{center}
\caption{Tensor decomposition of 4D tensor. While CP decomposition factorizes the 4D tensor into a summation of outer products of vector factors, MM decomposition factorizes the tensor into a summation of outer products of matrix factors(Eqn.\ref{eq:10}). }
\label{fig:MMdecomposition}
\end{figure}

We approximate the geometry grid and the appearance grid $\mathbf{G}_\sigma, \mathbf{G}_c$ with factorized tensors (Fig. \ref{fig:MMdecomposition}). The geometry grid is a 4D tensor $\mathbf{G}_{\sigma}\in \mathbb{R}^{I \times J \times K \times N}$ and the appearance grid is a 5D tensor $\mathbf{G}_{c}\in \mathbb{R}^{I \times J \times K \times N \times P}$. $I$, $J$, $K$, and $N$ correspond to the resolution of the feature grid along the $X$, $Y$,$Z$, and time axes respectively, and $P$ is the number of feature channels.

We decompose the dynamic radiance field with CP decomposition and Matrix-Matrix(MM) decomposition. Classic CP decomposition is directly applied to $\mathbf{G}_\sigma, \mathbf{G}_c$, and its decomposition equations are simply extended versions of Eqn.\ref{eq:4}, \ref{eq:5} with an additional vector corresponding to the time axis.

To reduce the number of components required to model a scene, we newly propose Matrix-Matrix(MM) decomposition, which only utilizes matrix factors.
Given a 4D tensor  $\mathbf{T}\in \mathbb{R}^{I \times J \times K \times N}$, MM decomposition can be expressed as:

\begin{align}\label{eq:10}
    \mathbf{T} = \sum_{r=1}^{R_1}M_{r}^{1,2}\circ M_{r}^{3,4} + \sum_{r=1}^{R_2}M_{r}^{1,3}\circ M_{r}^{2,4}+ \sum_{r=1}^{R_3}M_{r}^{2,3}\circ M_{r}^{1,4}
\end{align}

\begin{align*}
\text{where} \,\, M_{r}^{1,2} \in \mathbb{R}^{I \times J},  M_{r}^{1,3} \in \mathbb{R}^{I \times K}, M_{r}^{1,4} \in \mathbb{R}^{I \times N}, \\ M_{r}^{2,3} \in \mathbb{R}^{J \times K}, M_{r}^{2,4} \in \mathbb{R}^{J \times N}, M_{r}^{3,4} \in \mathbb{R}^{K \times N} 
\end{align*}

Each matrix corresponds to two modes of four modes ($X$, $Y$, $Z$, and time axes). While CP decomposition represents every mode with separate vectors,  we combine every two modes and represent them by matrices. Even though this relaxes the compactness of the model, it can express complex scenes with a smaller number of components compared to CP decomposition. Still memory complexity is decreased from  $\mathit{O}(n^4)$ to $\mathit{O}(n^2)$ compared to naive 4D grid representation.

With MM decomposition, the 4D geometry grid  and the 5D appearance grid $\mathbf{G}_\sigma, \mathbf{G}_c$ are factorized as follows:
\begin{align}\label{eq:11}
    \mathbf{G}_\sigma = \sum_{r=1}^{R_\sigma}M_{\sigma,r}^{X, Y}\circ M_{\sigma,r}^{Z, T} + M_{\sigma,r}^{X, Z}\circ M_{\sigma,r}^{Y, T} + M_{\sigma,r}^{Y, Z}\circ M_{\sigma,r}^{X, T}
\end{align}

\begin{align}\label{eq:12}
    \mathbf{G}_c = \sum_{r=1}^{R_c}M_{c,r}^{X, Y}\circ M_{c,r}^{Z, T}\circ b_{3r-2} + M_{c,r}^{X, Z}\circ M_{c,r}^{Y, T}\circ b_{3r-1} + M_{\sigma,r}^{Y, Z}\circ M_{c,r}^{X, T}\circ b_{3r}
\end{align}

With MM decomposition, the dynamic tensorial radiance field is factorized into $6R_c + 6R_\sigma$ matrices and $3R_c$ vectors.
We choose the value of $6R_c, 6R_\sigma$ smaller than any of $I\times J, I \times K, I \times N, J \times K, J \times N, N \times K$ which leads our representation still compact.

\subsection{Smoothing Regularization} \label{Smoothing regularization}

\begin{figure}[h]
\begin{center}

\centerline{\includegraphics[width=5cm]{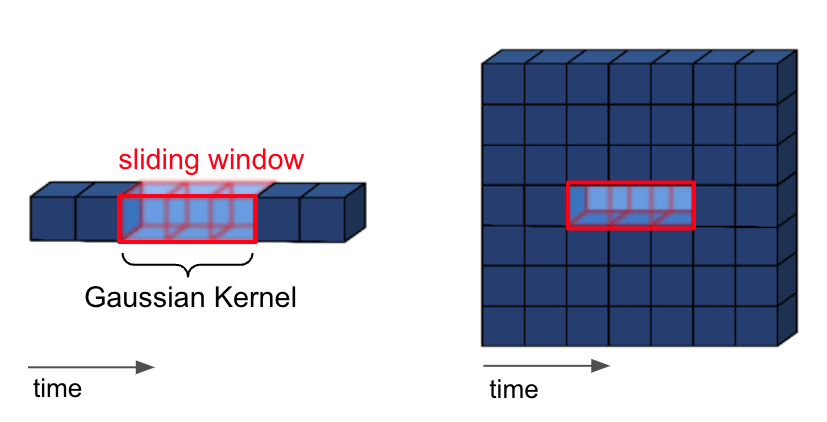}}

\end{center}
\caption{Smoothing regularization for models with CP decomposition(Left, Eqn.\ref{eq:13}) and MM decomposition(Right, Eqn.\ref{eq:14}). Smoothing regularization is applied while moving the predefined size of the window along the vector factor or the specific mode of matrix factor corresponding to the time axis.}
\label{fig: smoothing regularization}
\end{figure}

Simply extending the time dimension is not enough to produce a satisfactory result, and it is shown in Fig.\ref{fig:ta_ablation}. We must consider temporal dependency, which is the relation between the features at different times. We exploit the observation when the scene moves continuously over time; the appearance at t is most similar to that at t-1 and t+1. This means the features in the 4D grid should also be most close to those adjacent along the time axis.

To reflect temporal dependency, we introduce the regularization term(Eqn.\ref{eq:13}, \ref{eq:14}). We apply the regularization along time-related factors for both our models with CP decomposition and MM decomposition(Fig .\ref{fig: smoothing regularization}). 

\begin{align}\label{eq:13}
    L_{smooth} = \sum_{r=1}^{R}\sum_{v \in [{v_r^T}]}^{}\sum_{i_t=0}^{N}\left\|e_{i_t}^{v} - \sum_{i_w \in \mathit{window}(i_t, S)}^{}w(i_t, i_w)e_{i_w}^{v}\right\|^2
\end{align}
\begin{align}\label{eq:14}
    L_{smooth} = \sum_{r=1}^{R}\sum_{m\in [M_{r}^{X,T}, M_{r}^{Y,T}, M_{r}^{Z,T}]}\sum_{l=0}^{L}\sum_{i_t=0}^{N}\left\|e_{l,i_t}^{m} - \sum_{i_w \in \mathit{window}(i_t, S)}^{}w(i_t, i_w)e_{l,i_w}^{m}\right\|^2
\end{align}

$R$ is the value of $R_\sigma$ or $R_c$, which is the number of low-rank tensor components of the geometry grid or the appearance grid. $N$ is the resolution of the feature grid along the time axis, and $L$ indicates the resolution of other axes.  $i_w$ denotes index  which is close to $i_t$ in a $\mathit{window}(i_t, S)$ of size $S$. 
$w(i_t, i_w)$ is utilized to force some dependency between features at adjacent times. Inspired by TATD(\cite{tatd}), we choose the Gaussian kernel for the weight function $w(i_t, i_w)$ to reflect our observation that closer features along the time axis are more similar. The Gaussian kernel can be written as,

\begin{align}\label{eq:15}
    w(i_t, i_w) = \frac{K(i_t, i_w)}{\sum_{i_w' \in \mathit{window}(i_t, S)}K(i_t, i_w')}, K(i_t, i_w) = exp(-(i_t - i_w)^2 / 2\sigma^2)
\end{align}

Where $\sigma$ denotes the degree of smoothing. We used window size $S = 3$ and $\sigma = 0.5$ for our experiments. The Gaussian kernel has an additional advantage that it does not have any parameters to optimize, so our model can concentrate on learning the decomposed factors.

\subsection{Overall Framework}

\begin{figure}[h]
\begin{center}

\centerline{\includegraphics[width=\textwidth]{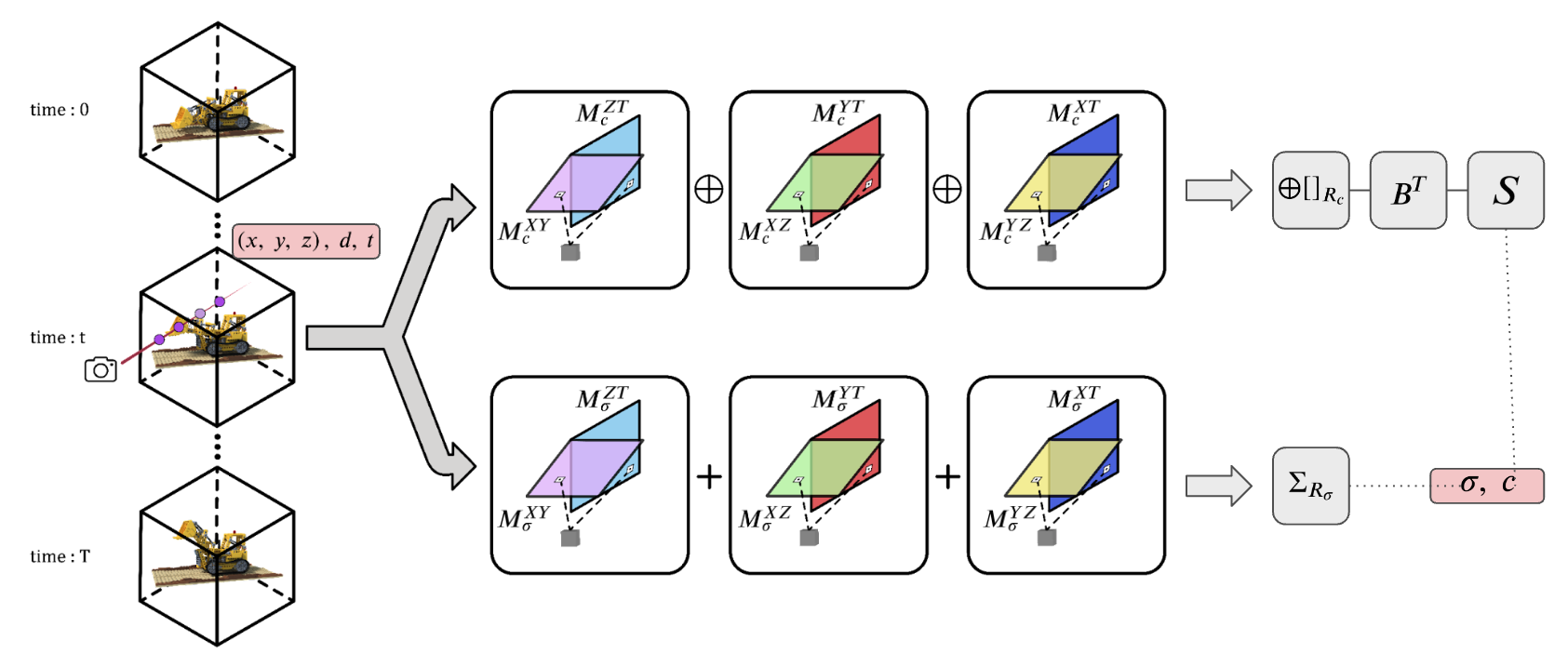}}

\end{center}
\caption{Overall framework of D-TensoRF with MM decomposition. }
\end{figure}

Our dynamic tensorial radiance field can be rewritten as:

\begin{equation}
    \begin{aligned}\label{eq:16}
         \sigma & = \mathbf{G}_\sigma(\textbf{x}, t) \\
         &= \sum_{r=1}^{R_\sigma}M_{\sigma,r}^{X, Y}(x,y)\circ M_{\sigma,r}^{Z, T}(z,t) + M_{\sigma,r}^{X, Z}(x,z)\circ M_{\sigma,r}^{Y, T}(y,t) + M_{\sigma,r}^{Y, Z}(y, z)\circ M_{\sigma,r}^{X, T}(x,t)
    \end{aligned}
\end{equation}
\begin{equation}
    \begin{aligned}\label{eq:17}
        \textbf{c} & = \mathbf{G}_c(\textbf{x}, t) \\
                &= \sum_{r=1}^{R_c}M_{c,r}^{X, Y}(x, y)\circ M_{c,r}^{Z, T}(z, t)\circ b_{3r-2} \\ 
                & \quad\quad\quad + M_{c,r}^{X, Z}(x, z)\circ M_{c,r}^{Y, T}(y, t)\circ b_{3r-1} + M_{\sigma,r}^{Y, Z}(y,z)\circ M_{c,r}^{X, T}(x, t)\circ b_{3r} \\
                &= \textbf{B}(\oplus[(M_{c,r}^{X, Y}(x, y)\circ M_{c,r}^{Z, T}(z, t)) \\
                & \quad\quad\quad \oplus (M_{c,r}^{X, Z}(x, z)\circ M_{c,r}^{Y, T}(y, t)) \oplus (M_{\sigma,r}^{Y, Z}(y,z)\circ M_{c,r}^{X, T}(x, t))]_{r=1,...,R_c})
    \end{aligned}
\end{equation}

Matrix $B$ is constructed by stacking all $b_r$ vectors. $\oplus$ is the concatenation operator, and it piles up all scalar values, which is a result of multiplying values from two matrices.
Stacked scalar values construct a vector whose dimension is $3R_c$.
After we obtain volume density $\sigma$ and view-dependent color $\textbf{c}$, we use differentiable volume rendering of NeRF(Eqn. \ref{eq:1}) to render images. 

Finally, we optimize our D-TensoRF by minimizing L2 rendering loss with 2D image supervision similar to Eqn. \ref{eq:2} via gradient descent. We utilize smoothing regularization to reflect temporal dependency. We also use an L1 regularization on our vector and matrix factors to prevent overfitting.

\subsection{Implementation Details}\label{implementation_details}

We implement our model based on the official implementation of TensoRF(\cite{TensoRF}). We use the appearance feature converting function $S$ as a small MLP, which consists of two fully-connected layers with 128-channel hidden layers. SH function also works as a function $S$, but the performance is slightly lower than the MLP, shown in Tab. \ref{tab:quantitative results}. Thus, we mostly choose to use the MLP. 

The number of components for the appearance grid $R_c$ is set to 3 times those for the geometry grid $R_\sigma$. 
We use the value of $N_t= 0.25 \times $ (number of training images) as the dimension of the feature grid along the time axis. 
We upsample the vector and matrix factors linearly and bilinearly at training steps 2000, 3000, 4000, 5500, and 7000. We set the initial resolution of the feature grid as $64^3 \times N_t$. We do not upsample the feature grid along time axes. 

In addition, we utilize a binary occupancy mask grid calculated at training step 2000 to make a more accurate bounding box that prevents unnecessary computation and precise modeling. We train D-TensoRF using a single GeForce RTX 3090 GPU. For other configurations, such as optimizer, learning rate, or batch size, we stick to the detail of the original TensoRF implementation. 

\section{Experiments}

\subsection{Experiments Setup}
We evaluate D-TensoRF on the extended Synthetic Nerf dataset provided by D-NeRF (\cite{pumarola2020d}). The dataset consists of eight scenes which include the motions of the objects. Each scene has $50 \sim 200$ training images and 20 test images. Images of each scene are obtained at different times. We train images at $400 \times 400$ pixels.

We also train TensoRF models with CP and VM decomposition for comparison with D-TensoRF models. We define the number of voxels in the feature grid of $64^3$ at the beginning of training. The number of final voxels is $150^3$ for TensoRF CP decomposition and $100^3$ for TensoRF-VM decomposition.

We conduct a quantitative and qualitative evaluation with various existing methods. NeRF (\cite{mildenhall2021nerf}), DirectVoxGo (\cite{Directvoxgo}), Plenoxels (\cite{plenoxel}), and TensoRF (\cite{TensoRF}) are the methods for reconstructing 3D static scenes. DirectVoxGo, Plenoxels, and TensoRF use explicit data structures to accelerate the training speed. We compare our D-TensoRFs with these methods to show that ours are suitable to model dynamic scenes. T-NeRF (\cite{pumarola2020d}) , D-NeRF (\cite{pumarola2020d}), TiNeuVox-S (\cite{tineuvox}), and TiNeuVox-B (\cite{tineuvox}) are targetted to dynamic scnenes. T-NeRF directly gives time information by extending an additional input dimension to the radiance field. D-Nerf, TiNeuVox-S, and TiNeuVox-B use an additional deformation field. Especially, TiNeuVox-S and TiNeuVox-B take advantage of time-aware neural voxels to reduce the training time. 

We use three evaluation metrics; Peak signal-to-noise ratio (PSNR), structural similarity (SSIM) (\cite{SSIM}), and learned perceptual image patch similarity (LPIPS)(\cite{LPIPS}). 
PSNR is the ratio between the maximum possible value of a signal and the power of distorting noise. For our experiments, It is defined by the maximum possible pixel value of the image and the mean squared error(MSE) between the rendered result and the ground truth. When PSNR is high, MSE is small, and the image is high-quality. SSIM measures the similarity between two images. SSIM is calculated in terms of luminance, contrast, and structure. The image is considered to be similar to the ground truth when the SSIM score is high. LPIPS estimates the perceptual similarity between two images. It evaluates the distance between images using the features extracted from the pre-trained deep network. Lower LPIPS means more perceptually similar.

In addition, we present an analysis of D-TensoRF models with different numbers of components and different numbers of voxels (the grid resolution). An ablation study is conducted on smoothing regularization to verify the importance of reflecting temporal dependency.

\subsection{Quantitative Evaluation}

\begin{table}[]
\begin{center}
    
\caption{ Evaluation scores of baseline methods are taken from their papers whenever available. Training time is estimated on a single RTX 3090 GPU. }
\label{tab:quantitative results}

\begin{threeparttable}
\resizebox{\textwidth}{!}{

\begin{tabular}{lccccc}
\multicolumn{1}{c}{\textbf{Method}} & \textbf{Time$\downarrow$} & \textbf{Size(MB)$\downarrow$} & \textbf{PSNR$\uparrow$} & \textbf{SSIM$\uparrow$} & \textbf{LPIPS$\downarrow$} \\ \hline
\multicolumn{1}{l|}{NeRF (\cite{mildenhall2021nerf})} & $\sim$hours & \multicolumn{1}{c|}{5} & 19.00 & 0.87 & 0.18 \\
\multicolumn{1}{l|}{DirectVoxGo (\cite{Directvoxgo})} & 5 mins & \multicolumn{1}{c|}{205} & 18.61 & 0.85 & 0.17 \\
\multicolumn{1}{l|}{Plenoxels (\cite{plenoxel})} & 6 mins & \multicolumn{1}{c|}{717} & 20.24 & 0.87 & 0.16 \\
\multicolumn{1}{l|}{TensoRF-CP384 (\cite{TensoRF})} & 14 mins & \multicolumn{1}{c|}{1} & 19.82 & 0.89 & 0.17 \\
\multicolumn{1}{l|}{TensoRF-VM192 (\cite{TensoRF})} & 9 mins & \multicolumn{1}{c|}{8} & 19.68 & 0.88 & 0.17 \\ \hdashline
\multicolumn{1}{l|}{T-NeRF (\cite{pumarola2020d})} & $\sim$hours & \multicolumn{1}{c|}{-} & 29.51 & 0.96 & 0.08 \\
\multicolumn{1}{l|}{D-NeRF (\cite{pumarola2020d})} & 20 hours & \multicolumn{1}{c|}{4} & 30.50 & 0.95 & 0.07 \\
\multicolumn{1}{l|}{TiNeuVox-S (\cite{tineuvox})} & 8 mins & \multicolumn{1}{c|}{8} & 30.75 & 0.96 & 0.07 \\
\multicolumn{1}{l|}{TiNeuVox-B (\cite{tineuvox})} & 28mins & \multicolumn{1}{c|}{48} & 32.67 & 0.97 & 0.04 \\ \hdashline
\multicolumn{1}{l|}{Ours-CP768 ($150^3 \times N_t$, 60k steps)} & 39 mins & \multicolumn{1}{c|}{1.8} & 30.68 & 0.96 & 0.03  \\
\multicolumn{1}{l|}{Ours-MM192 ($100^3 \times N_t$, 30k steps)} & 8 mins & \multicolumn{1}{c|}{10.8}   &29.74  &0.96  &0.04 \\
\multicolumn{1}{l|}{Ours-MM192 ($100^3 \times N_t$, 60k steps)} & 16 mins & \multicolumn{1}{c|}{10.8}   &29.94 &0.96 & 0.03 \\
\multicolumn{1}{l|}{Ours-CP768-SH ($150^3 \times N_t$, 60k steps)} & 42 mins & \multicolumn{1}{c|}{1.6}  &29.52  &0.95  &0.04  \\
\multicolumn{1}{l|}{Ours-MM192-SH ($100^3 \times N_t$, 60k steps)} & 16 mins & \multicolumn{1}{c|}{10.6}   &27.7 &0.94 &0.06 
\end{tabular}

}
\end{threeparttable}

\end{center}
\end{table}

In Tab. \ref{tab:quantitative results}, we show the quantitative results of D-TensoRF-CP, D-TensoRF-MM, and the existing methods.
As NeRF and TensoRF aim at modeling static scenes, it is shown that they do not correctly model dynamic scenes in all evaluation metrics.

T-NeRF, a NeRF method for modeling dynamic scenes, can be considered to be a similar approach to ours, but D-TensoRF-CP and D-TensoRF-MM outperformed T-Nerf in all evaluation metrics. D-Nerf, TiNeuVox-S, and TiNeuVox-B are the methods that use an deformation network and show good evaluation scores. Compared to them, D-TensoRF-CP or D-TensoRF-MM have slightly lower PSNR scores, but SSIM and LPIPS show almost similar or better scores. Our models consistently stand out in the LPIPS evaluation metric, although the PSNR score is slightly lower than the best score.

We also report the per-scene quantitative comparisons in Tab.\ref{tab:per-scene}. D-TensoRF-CP and D-TensoRF-MM show satisfactory LPIPS scores in all scenes. In particular, in scenes where the motion between images at adjacent times is not very large, ours have better PSNR and SSIM scores than baseline methods.

Regarding training speed, because T-Nerf and D-Nerf use only MLP purely, they are very slow (about 20 hours), whereas TiNeuVox shows a faster speed by utilizing explicit data structure. In particular, the TiNeuVox-S model shows promising results with an 8-minute training speed. Our D-TensoRF-MM with 30k iterations obtains a better LPIPS score with the same 8-minute training speed. We can improve further by training up to 60k iterations, which took 16 minutes. D-TensoRF-CP takes a longer training time than D-TensoRF-MM at 39 minutes, but still, we can produce 
high evaluation scores in a training time of less than 1 hour.

Our model requires a significantly smaller memory footprint than all previous methods modeling dynamic scenes. The most compact model among previous methods, D-NeRF, required 4MB, but our D-TensoRF-CP required only 1.8MB, even though we used an explicit feature grid for fast training speed. D-TensoRF-MM requires 10.8 MB, slightly larger than the TiNeuVox-S and much more compact than TiNeuVox-B.

We also use the SH function to convert features into color and density, but it shows lower scores in all evaluation metrics compared to when the MLP is used. Still, LPIPS scores are similar to those of D-NerF and TiNeuVox-S.

\subsubsection{Revisiting PSNR score}

\begin{figure}[h]
\begin{center}

\centerline{\includegraphics[width=8cm]{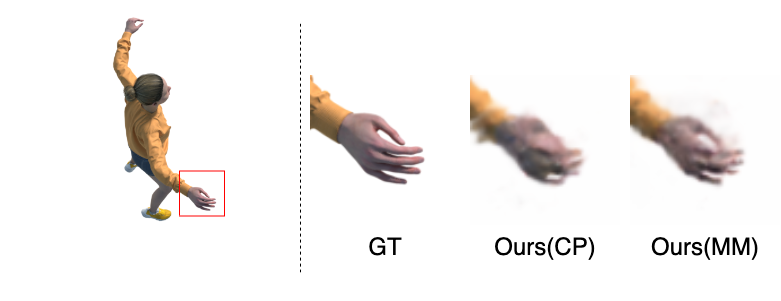}}

\end{center}
\caption{Addtional rendering results of D-TensoRF with CP decomposition(Left) and MM decomposition(Right). D-TensoRF-CP tends to produce blurry results compared to D-TensoRF-MM.}
\label{fig: CPvsMM}
\end{figure}

We can observe that PSNR prefers blurry images to sharp images in Fig. \ref{fig: CPvsMM}, \ref{Qualitive Evaluation}. In Fig. \ref{fig: CPvsMM}, even the rendering result of D-TensoRF-MM produces sharper images than D-TensoRF-CP, but the PSNR score is slightly lower. This is because PSNR is calculated based on pixel-level mean square error, so it is too sensitive to small shifts. Therefore, it cannot guarantee that the rendering quality of a model with a high PSNR score is satisfactory for human eyes. This phenomenon is also pointed out by \cite{hypernerf}.

\subsection{Qualitative Evaluation} \label{Qualitive Evaluation}

\begin{figure}[h]
\begin{center}

\centerline{\includegraphics[width=\textwidth]{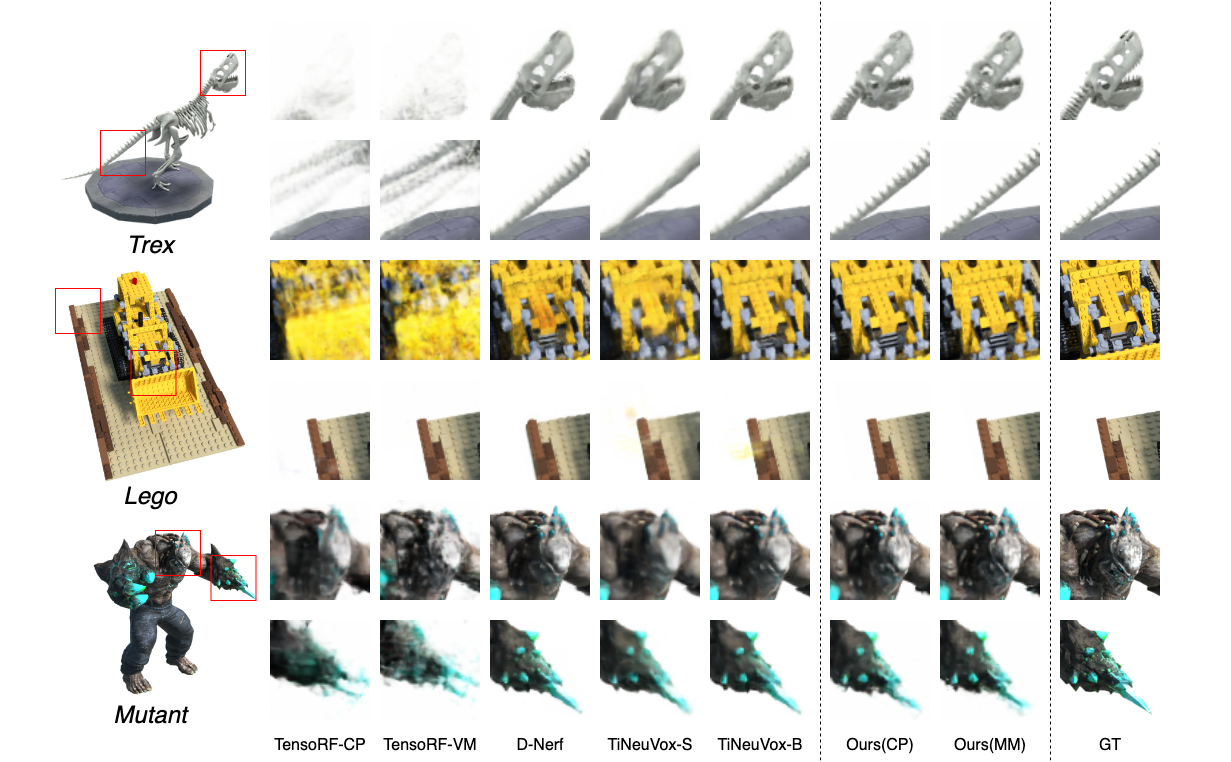}}

\end{center}
\caption{Qualitative results of D-TensoRF-CP, D-TensoRF-MM and baseline methods on the extended Synthetic NeRF scenes provided by D-NeRF(\cite{pumarola2020d}).}
\label{fig: qualitative}
\end{figure}

We provide qualitative results of D-TensoRF-CP, D-TensoRF-MM, and the comparison methods in Fig.\ref{Qualitive Evaluation}.
The two decomposition methods of TensoRF fail to model a dynamic scene. In particular, since the movement is not reflected at all, they give results that seem to indicate that some training images overlapped.

D-NeRF, Tineuvox-s, and Tineuvox-B appear to predict motion well. However, there is a problem of misprediction, which is a disadvantage of using a deformation network. In the Lego rendering results, D-NeRF shows an unrelated brown color where it should be yellow. Moreover, yellow appears where nothing should be in the results of Tineuvox-S and Tineuvox-B. Our methods, D-TensoRF-CP and D-TensoR-MM, do not have a misprediction problem because we do not use a deformation network.

Also, the overall results of the existing methods are blurred. In the rendered images of the T-rex, Tineuvox-S and Tineuvox-B can not express the pointy part. D-NeRF shows a sharper image than Tineuvox-S and Tineuvox-B, but the image is still more blurred than the results of D-TensoRF-CP and D-TensoRF-MM.

Our methods produce the sharpest image when there is a smooth motion. However, in the case of the mutant's hand, the rendering result seems to be smudged compared to the ground truth. The result is similar to Tineuvox-B and Tineuvox-S and a little more smudged than D-NeRF. We presume this is because the smoothing regularization using the Gaussian kernel cannot perfectly model the temporal dependency when there is a large motion between the scenes in adjacent times.

We further demonstrate that D-TensoRF-CP and D-TensoRF-MM synthesize high-quality images of every scene at different times and unseen viewing directions in Fig. \ref{fig: novelviewcp}, \ref{fig: novelviewmm}, \ref{fig: noveltimecp}, and \ref{fig: noveltimemm}.

\subsection{Analysis of different D-TensoRF models}
\begin{table}[h]

\begin{center}

\caption{Evaluation scores of D-TensoRF-CP and D-TensoRF-MM with different number of components and final voxels.}
\label{tab:qualitive_numvoxel}

{\small
\begin{threeparttable}
\resizebox{\textwidth}{!}{
\begin{tabular}{c|c|ccc|ccc|ccc}
\hline
\multirow{2}{*}{} & \multirow{2}{*}{\#Comp} & \multicolumn{3}{c|}{$64^3 \times N_t$} & \multicolumn{3}{c|}{$100^3 \times N_t$} & \multicolumn{3}{c}{$150^3 \times N_t$} \\
 &  & PSNR$\uparrow$ & SSIM$\uparrow$ & LPIPS$\downarrow$ & PSNR$\uparrow$ & SSIM$\uparrow$ & LPIPS$\downarrow$ & PSNR$\uparrow$ & SSIM$\uparrow$ & LPIPS$\downarrow$ \\ \hline
\multirow{3}{*}{D-TensoRF-CP} & 192 & 16.38 & 0.883 & 0.241 & 16.45 & 0.884 & 0.243 & 16.46 & 0.885 & 0.241 \\
 & 384 & 28.95 & 0.951 & 0.052 & 29.06 & 0.950 & 0.046 & 29.00 & 0.953 & 0.045 \\
 & 768 & 30.41 & 0.961 & 0.038 & 29.64 & 0.963 & 0.033 & 30.68 & 0.963 & 0.031 \\ \hline
\multirow{3}{*}{D-TensoRF-MM} & 96 & 29.63 & 0.956 & 0.043 & 29.55 & 0.958 & 0.040 & 29.26 & 0.954 & 0.043 \\
 & 192 & 29.97 & 0.961 & 0.035 & 29.94 & 0.960 & 0.034 & 29.71 & 0.958 & 0.034 \\
 & 384 & 29.88 & 0.963 & 0.030 & 29.93 & 0.963 & 0.030 & 29.66 & 0.960 & 0.031 \\ \hline
\end{tabular}
}
\end{threeparttable}
}
\end{center}
\end{table}

\begin{table}[h]
\begin{center}
\caption{Comparison of training speed and memory footprint in various settings. Training speed and memory footprint increases while the number of components and voxels increases.}
\label{tab:my-table}
\begin{threeparttable}
\resizebox{\textwidth}{!}{
\begin{tabular}{c|c|cc|cc|cc}
\hline
\multirow{2}{*}{} & \multirow{2}{*}{\#Comp} & \multicolumn{2}{c|}{$64^3 \times N_t$} & \multicolumn{2}{c|}{$100^3 \times N_t$} & \multicolumn{2}{c}{$150^3 \times N_t$} \\
 &  & Time$\downarrow$ & Size(MB)$\downarrow$ & Time$\downarrow$ & Size(MB)$\downarrow$ & Time$\downarrow$ & Size(MB)$\downarrow$ \\ \hline
\multirow{3}{*}{D-TensoRF-CP} & 192 & 19:26 & 0.35 & 24:24 & 0.44 & 31:50 & 0.55 \\
 & 384 & 18:10 & 0.58 & 20:58 & 0.74 & 26:51 & 0.99 \\
 & 768 & 23:34 & 0.95 & 29:52 & 1.26 & 38:40 & 1.75 \\ \hline
\multirow{3}{*}{D-TensoRF-MM} & 96 & 14:46 & 2.64 & 15:42 & 5.46 & 18:21 & 11.13 \\
 & 192 & 15:01 & 5.14 & 15:55 & 10.83 & 19:53 & 22.21 \\
 & 384 & 15:39 & 10.10 & 16:58 & 21.33 & 23:15 & 43.89 \\ \hline
\end{tabular}
}
\end{threeparttable}
\end{center}
\end{table}

We conduct an extensive evaluation while doubling the number of components from 192 to 768 for D-TensoRF-CP and doubling the number of components from 96 to 384 for D-TensoRF-MM. Furthermore, we perform experiments when the number of final voxels is $64^3 \times N_t$, $100^3\times N_t$, and $150^3\times N_t$. In Tab.\ref{tab:qualitive_numvoxel}, we show the qualititative comparisons between different D-TensoRFs.
 
In the case of D-TensoRF-CP, 192 components are not enough to model a dynamic scene. This is because only a little information can be contained per component due to the high compactness of the CP decomposition. The modeling performance comes out well when more than 384 components are used. D-TensoRF-MM shows high-quality results even with 96 components. The performances of both D-TensoRF-CP and D-TensoRF-MM tend to improve as the number of components grows.

However, the evaluation scores do not improve as the number of voxels increases.
The number of points according to the coordinate $(\textbf{x}, t)$ included in one voxel of the feature grid increases when the grid resolution is too small. Thus, much information is compressed into one feature while optimizing the feature grid. As the expressiveness of the model reduces, the synthesized image is not sharp as a result. 
On the other hand, when the number of voxels is too large, only a few points corresponding to the coordinate $(\textbf{x},t)$ are included in one voxel. The feature of non-moving parts acquires additional information from the features in adjacent times through smoothing regularization. However, the feature corresponding to moving parts even cannot obtain sufficient information from the features in adjacent times. Consequently, while the model with superabundant voxels can express non-moving parts in detail, the model renders a smudged image of moving parts. 

In Fig.\ref{fig:num_voxel}, we show the rendered results when the number of voxels is $64^3 \times N_t $ and $150^3 \times N_t $. The synthesized images are blurry with $64^3 \times N_t $ voxels. The results with $150^3 \times N_t $ voxels can express the details sharp, but some appear to be smudged.
We, therefore, select  $150^3 \times N_t$ for D-TensoRF-CP, $100^3 \times N_t$ for D-TensoRF-MM as the number of final voxels which achieves the best rendering quality. 

\subsection{Ablation study on smoothing regularization}
\begin{figure}[h]
\begin{center}

\centerline{\includegraphics[width=6cm]{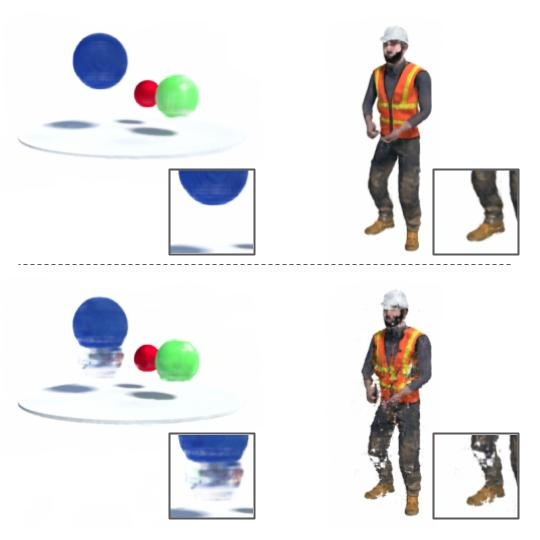}}

\end{center}
\caption{Ablation study on smoothing regularization in terms of rendering quality. The above images are the results when we apply the smoothing regularization. The below images are the results of not applying the smoothing regularization.}
\label{fig:ta_ablation}
\end{figure}

An ablation study on smoothing regularization is also conducted. In Tab.\ref{tab:ablation_smooth}, evaluation metric scores increased by applying smoothing regularization. In Fig. \ref{fig:ta_ablation}, it can be seen that the quality significantly increases when the temporal dependency is reflected through smoothing regularization.

\begin{table}[]
\begin{center}
\caption{Ablation study on smoothing regularization in terms of evaluation scores.}
\label{tab:ablation_smooth}
\begin{tabular}{c|ccc}
Method & PSNR$\uparrow$ & SSIM$\uparrow$ & LPIPS$\downarrow$ \\ \hline
D-TensoRF-CP768($150^3 \times N_t$, w/o smoothing regularization) &28.91  &0.95  &0.04  \\
D-TensoRF-CP768($150^3\times N_t$, w smoothing regularization) &30.68  &0.96 &0.03   \\ \hdashline
D-TensoRF-MM192($100^3\times N_t$, w/o smoothing regularization) &27.72  &0.95  &0.04  \\
D-TensoRF-MM192($100^3\times N_t$, w smoothing regularization) &29.94  &0.96  & 0.03 \\ 
\end{tabular}

\end{center}
\end{table}

\section{Limitations and Future Work}
 
We use a smoothing regularization using a Gaussian kernel based on the observation that the scene at time $t$ most closely resembles the scenes at $t-1$, $t+1$. However, as shown in section \ref{Qualitive Evaluation}, it fails if there are large movements between adjacent scenes in the training dataset. Including these cases, our method could be further developed by applying a weight function other than the Gaussian kernel that more accurately reflects temporal dependencies. 

Also, D-TensoRF can be applied only to bounded scenes similar to TensoRF. We suggest modeling the background and objects with motion separately can be the one way to resolve this limitation. Or, other approaches could possibly enable our methods to apply to unbounded scenes, we leave this as future work.

\section{Conclusion}

We present a method for synthesizing a high-quality scene in a novel time and viewing direction by modeling a dynamic scene. We decompose the dynamic radiance field into low-rank tensor components using the classic CP decomposition, and the newly proposed MM decomposition. Through this, the training time could be significantly reduced, as could the required memory footprint compared to existing methods.

In addition, by reflecting the temporal dependency, the motion is implicitly modeled using smoothing regularization. As a result, we can model the dynamic scene efficiently without using a deformation network.

D-TensoRF with CP decomposition and MM decomposition obtains competitive scores in all evaluation metrics. Especially, we achieve the state-of-the-art LPIPS scores, which reflects the perceptual quality. Our method have sharper images compared to existing methods in terms of quality. In addition, our method simultaneously achieve fast training time and a small memory footprint.

Based on 60k training steps, the training time is significantly reduced to 40 minutes for D-TensoRF with CP decomposition and to 16 minutes for MM decomposition. Also, the memory footprint is highly compact at 1.8MB for CP decomposition and at 10.8MB for MM decomposition.

\bibliography{iclr2023_conference}
\bibliographystyle{iclr2023_conference}

\appendix
\section{Appendix}


\begin{table}
\caption{Per-scene quantitative evaluation on the extended Synthetic Nerf dataset provided by D-NeRF (\cite{pumarola2020d}).}
\label{tab:per-scene}
\begin{center}
    
{\tiny
\begin{tabular}{l|lll|llllll}
\hline
\multicolumn{1}{c|}{\multirow{2}{*}{Method}} & \multicolumn{3}{c|}{\textbf{Hell Warrior}} & \multicolumn{3}{c|}{\textbf{Mutant}} & \multicolumn{3}{c}{\textbf{Hook}} \\
\multicolumn{1}{c|}{} & \multicolumn{1}{c}{PSNR$\uparrow$} & \multicolumn{1}{c}{SSIM$\uparrow$} & \multicolumn{1}{c|}{LPIPS$\downarrow$} & \multicolumn{1}{c}{PSNR$\uparrow$} & \multicolumn{1}{c}{SSIM$\uparrow$} & \multicolumn{1}{c|}{LPIPS$\downarrow$} & \multicolumn{1}{c}{PSNR$\uparrow$} & \multicolumn{1}{c}{SSIM$\uparrow$} & \multicolumn{1}{c}{LPIPS$\downarrow$} \\ \hline
NeRF (\cite{mildenhall2021nerf}) & 13.52 & 0.81 & 0.25 & 20.31 & 0.91 & \multicolumn{1}{l|}{0.09} & 16.65 & 0.84 & 0.19 \\
DirectVoxGO (\cite{Directvoxgo}) & 13.52 & 0.75 & 0.25 & 19.45 & 0.89 & \multicolumn{1}{l|}{0.12} & 16.16 & 0.80 & 0.21 \\
Plenoxels (\cite{plenoxel}) & 15.19 & 0.78 & 0.27 & 21.44 & 0.91 & \multicolumn{1}{l|}{0.09} & 17.90 & 0.81 & 0.21 \\
TensoRF-CP384 (\cite{TensoRF}) & 14.2 & 0.81 & 0.28 & 21.05 & 0.91 & \multicolumn{1}{l|}{0.11} & 17.83 & 0.84 & 0.2 \\
TensoRF-VM192 (\cite{TensoRF}) & 13.53 & 0.77 & 0.31 & 20.81 & 0.91 & \multicolumn{1}{l|}{0.11} & 17.27 & 0.83 & 0.2 \\
T-NeRF (\cite{pumarola2020d}) & 23.19 & 0.93 & 0.08 & 30.56 & 0.96 & \multicolumn{1}{l|}{0.04} & 27.21 & 0.94 & 0.06 \\
D-NeRF (\cite{pumarola2020d}) & 25.02 & 0.95 & 0.06 & 31.29 & 0.97 & \multicolumn{1}{l|}{0.02} & 29.25 & 0.96 & 0.11 \\
TiNeuVox-S (\cite{tineuvox}) & 27.00 & 0.95 & 0.09 & 31.09 & 0.96 & \multicolumn{1}{l|}{0.05} & 29.30 & 0.95 & 0.07 \\
TiNeuVox-B (\cite{tineuvox}) & 28.17 & 0.97 & 0.07 & 33.61 & 0.98 & \multicolumn{1}{l|}{0.03} & 31.45 & 0.97 & 0.05 \\
D-TensoRF-CP768 & 23.65 & 0.93 & 0.07 & 32.35 & 0.97 & \multicolumn{1}{l|}{0.02} & 27.73& 0.95 & 0.04  \\
D-TensoRF-MM192 & 22.59  & 0.92 &0.08  &32.63  &0.97  & \multicolumn{1}{l|}{0.02} & 27.54  &0.95  &0.04  \\ \hline
\multicolumn{1}{c|}{\multirow{2}{*}{Method}} & \multicolumn{3}{c|}{\textbf{Bouncing Balls}} & \multicolumn{3}{c|}{\textbf{Lego}} & \multicolumn{3}{c}{\textbf{T-Rex}} \\
\multicolumn{1}{c|}{} & \multicolumn{1}{c}{PSNR$\uparrow$} & \multicolumn{1}{c}{SSIM$\uparrow$} & \multicolumn{1}{c|}{LPIPS$\downarrow$} & \multicolumn{1}{c}{PSNR$\uparrow$} & \multicolumn{1}{c}{SSIM$\uparrow$} & \multicolumn{1}{c|}{LPIPS$\downarrow$} & \multicolumn{1}{c}{PSNR$\uparrow$} & \multicolumn{1}{c}{SSIM$\uparrow$} & \multicolumn{1}{c}{LPIPS$\downarrow$} \\ \hline
NeRF (\cite{mildenhall2021nerf}) & 20.26 & 0.91 & 0.20 & 20.30 & 0.79 & \multicolumn{1}{l|}{0.23} & 24.49 & 0.93 & 0.13 \\
DirectVoxGO (\cite{mildenhall2021nerf}) & 20.20 & 0.87 & 0.22 & 21.13 & 0.90 & \multicolumn{1}{l|}{0.10} & 23.27 & 0.92 & 0.09 \\
Plenoxels (\cite{plenoxel}) & 21.30 & 0.89 & 0.18 & 21.97 & 0.90 & \multicolumn{1}{l|}{0.11} & 25.18 & 0.93 & 0.08 \\
TensoRF-CP384 (\cite{TensoRF}) & 21.03 & 0.9 & 0.17 & 21.61 & 0.91 & \multicolumn{1}{l|}{0.09} & 25.25 & 0.94 & 0.08 \\
TensoRF-VM192 (\cite{TensoRF}) & 20.89 & 0.9 & 0.16 & 21.44 & 0.9 & \multicolumn{1}{l|}{0.1} & 25.1 & 0.94 & 0.08 \\
T-NeRF (\cite{pumarola2020d}) & 37.81 & 0.98 & 0.12 & 23.82 & 0.90 & \multicolumn{1}{l|}{0.15} & 30.19 & 0.96 & 0.13 \\
D-NeRF (\cite{pumarola2020d}) & 38.93 & 0.98 & 0.10 & 21.64 & 0.83 & \multicolumn{1}{l|}{0.16} & 31.75 & 0.97 & 0.03 \\
TiNeuVox-S (\cite{tineuvox}) & 39.05 & 0.99 & 0.06 & 24.35 & 0.88 & \multicolumn{1}{l|}{0.13} & 29.95 & 0.96 & 0.06 \\
TiNeuVox-B (\cite{tineuvox}) & 40.73 & 0.99 & 0.04 & 25.02 & 0.92 & \multicolumn{1}{l|}{0.07} & 32.70 & 0.98 & 0.03 \\
D-TensoRF-CP768 & 41.52 & 0.99 & 0.005 & 25.38 & 0.94 & \multicolumn{1}{l|}{0.03} & 31.23 & 0.97 & 0.03 \\
D-TensoRF-MM192 & 39.15  &0.99  &0.01  &25.30  &0.94 & \multicolumn{1}{l|}{0.03} &30.18  &0.97  &0.03  \\ \hline
\multicolumn{1}{c|}{\multirow{2}{*}{Method}} & \multicolumn{3}{c|}{\textbf{Stand Up}} & \multicolumn{3}{c}{\textbf{Jumping Jacks}} & \multicolumn{3}{c}{\textbf{}} \\
\multicolumn{1}{c|}{} & \multicolumn{1}{c}{PSNR$\uparrow$} & \multicolumn{1}{c}{SSIM$\uparrow$} & \multicolumn{1}{c|}{LPIPS$\downarrow$} & \multicolumn{1}{c}{PSNR$\uparrow$} & \multicolumn{1}{c}{SSIM$\uparrow$} & \multicolumn{1}{c}{LPIPS$\downarrow$} & \multicolumn{1}{c}{} & \multicolumn{1}{c}{} & \multicolumn{1}{c}{} \\ \cline{1-7}
NeRF (\cite{mildenhall2021nerf}) & 18.19 & 0.89 & 0.14 & 18.28 & 0.88 & 0.23 &  &  &  \\
DirectVoxGO (\cite{mildenhall2021nerf}) & 17.58 & 0.86 & 0.16 & 17.80 & 0.84 & 0.20 &  &  &  \\
Plenoxels (\cite{plenoxel}) & 18.76 & 0.87 & 0.15 & 20.18 & 0.86 & 0.19 &  &  &  \\
TensoRF-CP384 (\cite{TensoRF}) & 18.29 & 0.88 & 0.18 & 19.29 & 0.88 & 0.23 & & & \\
TensoRF-VM192 (\cite{TensoRF}) & 18.96 & 0.89 & 0.15 & 19.41 & 0.87 & 0.24 & & & \\
T-NeRF (\cite{pumarola2020d}) & 31.24 & 0.97 & 0.02 & 32.01 & 0.97 & 0.03 &  &  &  \\
D-NeRF (\cite{pumarola2020d}) & 32.79 & 0.98 & 0.02 & 32.80 & 0.98 & 0.03 &  &  &  \\
TiNeuVox-S (\cite{tineuvox}) & 32.89 & 0.98 & 0.03 & 32.33 & 0.97 & 0.04 &  &  &  \\
TiNeuVox-B (\cite{tineuvox}) & 35.43 & 0.99 & 0.03 & 34.23 & 0.98 & 0.03 &  &  &  \\
D-TensoRF-CP768 & 32.59 & 0.98 & 0.02 & 31.01 & 0.97 & 0.03 &  &  &  \\
D-TensoRF-MM192 & 32.00 & 0.97 & 0.02 & 30.13  & 0.97  & 0.04  &  &  &  \\ \cline{1-7}
\end{tabular}
}
\end{center}
\end{table}

\begin{figure}[h]
\begin{center}

\centerline{\includegraphics[width=6cm]{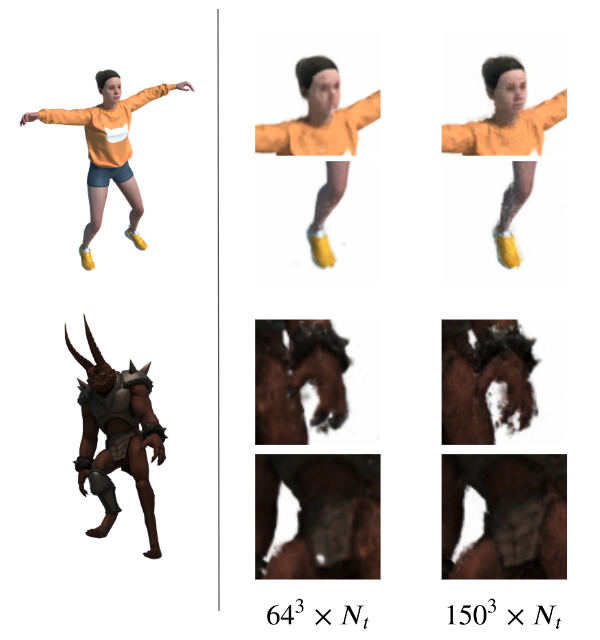}}

\end{center}
\caption{Synthesized images of D-TensoRF-MM192 with different numbers of voxels.}
\label{fig:num_voxel}
\end{figure}

\begin{figure}[h]
\begin{center}

\centerline{\includegraphics[width=12cm]{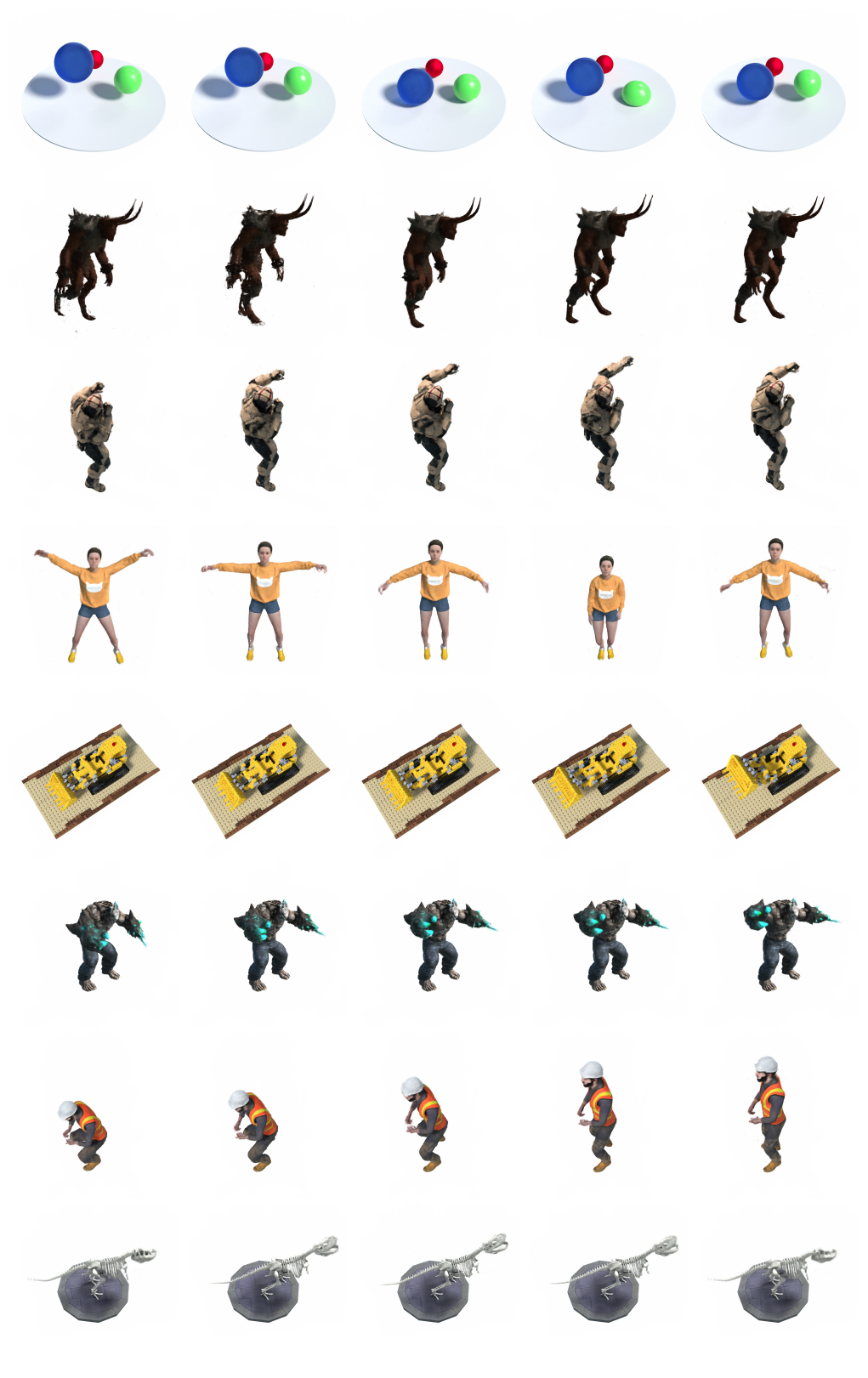}}

\end{center}
\caption{Rendering results of D-TensoRF-CP. Images of each scene are synthesized under different time conditions with the same viewing direction. The eight scenes are provided by D-NeRF 
 (\cite{pumarola2020d}).}
\label{fig: novelviewcp}
\end{figure}

\begin{figure}[h]
\begin{center}

\centerline{\includegraphics[width=12cm]{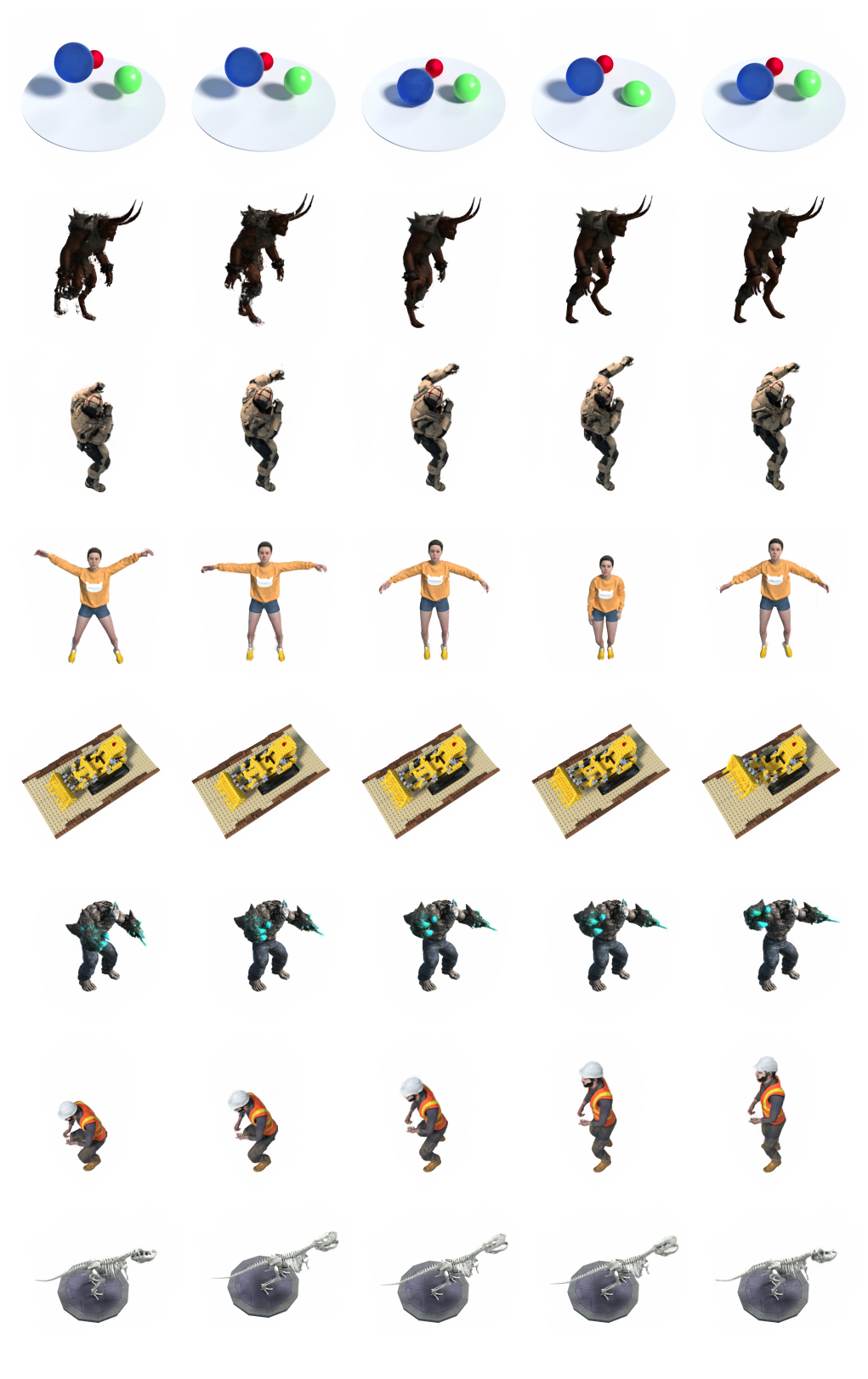}}

\end{center}
\caption{Rendering results of D-TensoRF-MM. Images of each scene are synthesized under different time conditions with the same viewing direction. The eight scenes are provided by D-NeRF (\cite{pumarola2020d}).}
\label{fig: novelviewmm}
\end{figure}

\begin{figure}[h]
\begin{center}

\centerline{\includegraphics[width=12cm]{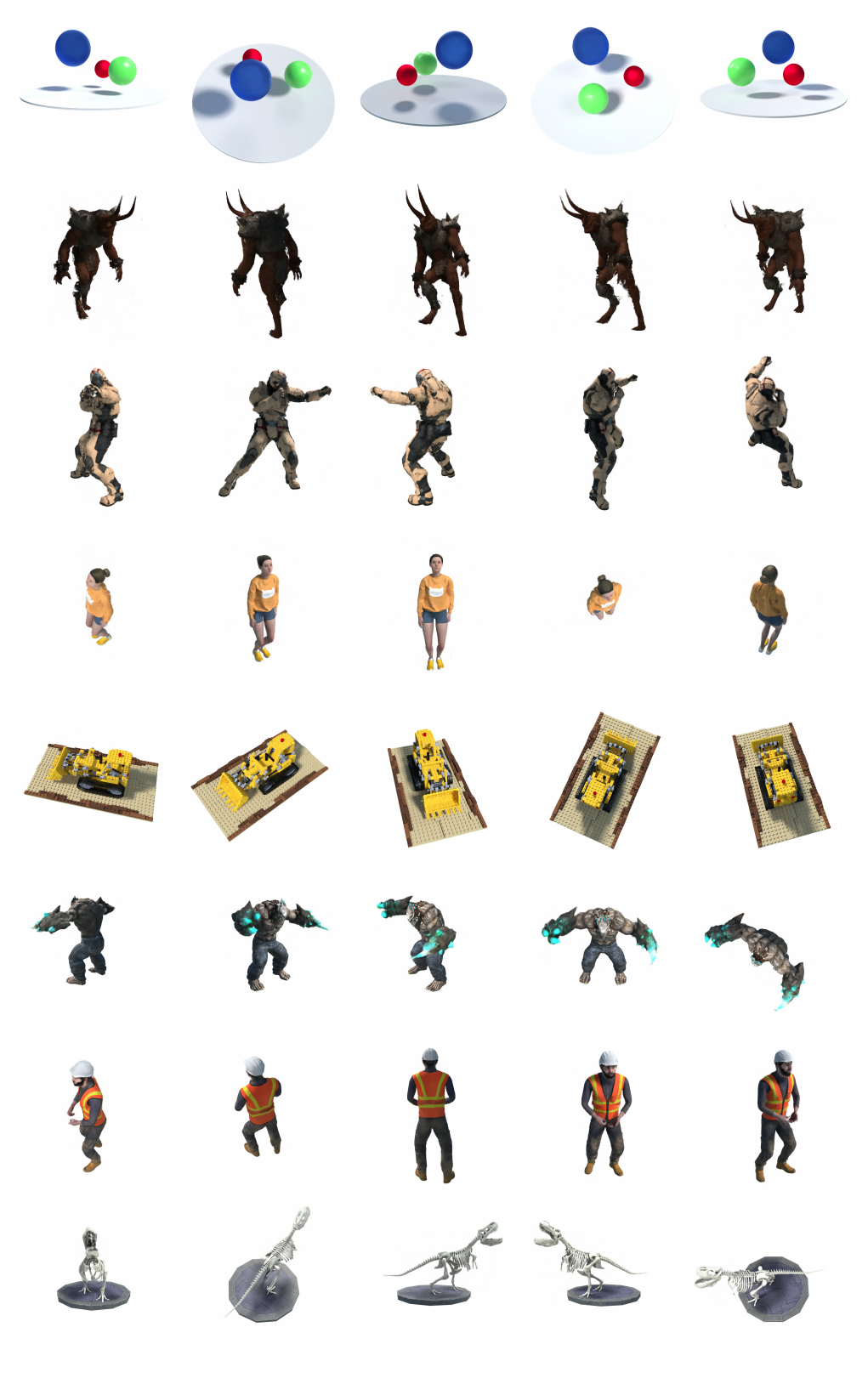}}

\end{center}
\caption{Rendering results of D-TensoRF-CP. Images of each scene are synthesized under different viewing directions with the same time condition. The eight scenes are provided by D-NeRF (\cite{pumarola2020d}).}
\label{fig: noveltimecp}
\end{figure}

\begin{figure}[h]
\begin{center}

\centerline{\includegraphics[width=12cm]{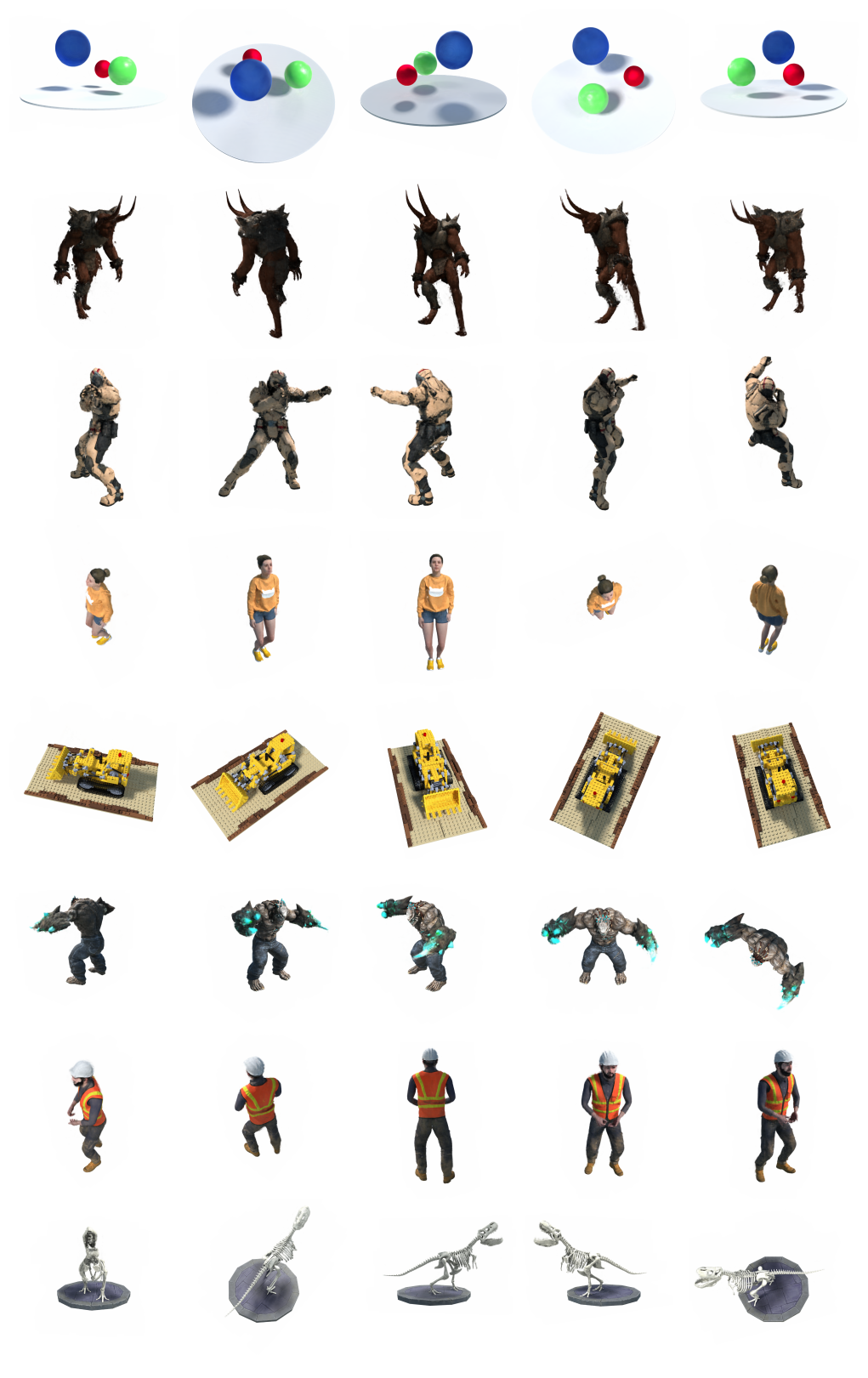}}

\end{center}
\caption{Rendering results of D-TensoRF-MM. Images of each scene are synthesized under different viewing directions with the same time condition. The eight scenes are provided by D-NeRF (\cite{pumarola2020d}).}
\label{fig: noveltimemm}
\end{figure}

\end{document}

%% file: math_commands.tex
\usepackage{amsmath,amsfonts,bm}

\def\eqref#1{equation~\ref{#1}}

\def\1{\bm{1}}

\DeclareMathAlphabet{\mathsfit}{\encodingdefault}{\sfdefault}{m}{sl}
\SetMathAlphabet{\mathsfit}{bold}{\encodingdefault}{\sfdefault}{bx}{n}

